\documentclass[3p,12pt,authoryear]{elsarticle}

\usepackage{helvet}  
\usepackage{courier}  
\usepackage{natbib}
\usepackage{bm} 
\usepackage{algorithm}
\usepackage{algorithmic}
\usepackage{amssymb}
\usepackage{amsmath}
\usepackage{cases}
\usepackage{booktabs}
\usepackage{array}
\usepackage{multirow}
\usepackage{bm} 
\usepackage{booktabs}
\usepackage{multirow}
\usepackage{amsmath}
\usepackage{amssymb}
\usepackage[dvipsnames]{xcolor}
\usepackage[most]{tcolorbox}
\usepackage{subcaption}
\usepackage{pifont}
\usepackage{framed}
\usepackage{listings}
\usepackage{rotating}
\usepackage{adjustbox}
\PassOptionsToPackage{table}{xcolor}
\definecolor{lightgray}{rgb}{0.87,0.87,0.87}
\definecolor{blue1}{rgb}{0.06,0.42,0.53}
\definecolor{red1}{rgb}{0.72,0.18,0.11}
\definecolor{mygreen}{rgb}{0,0.6,0}
\definecolor{myblue}{rgb}{0,0,0.6}
\definecolor{mycyan}{rgb}{0,0.6,0.6}
\definecolor{lightgreen}{rgb}{0.890, 0.941, 0.855}
\usepackage{colortbl}

\journal{}

\begin{document}

\begin{frontmatter}



\title{Uncertainty-Guided Self-Questioning and Answering for Video-Language Alignment} 

\author[a,b]{Jin Chen}
\author[a,b]{Kaijing Ma}
\author[c]{Haojian Huang}
\author[a]{Han Fang}
\author[a]{Hao Sun}
\author[d]{Mehdi Hosseinzadeh}
\author[e]{Zhe Liu}

\affiliation[a]{
organization={TeleAI},
city={Beijing},
postcode={100007},
country={China}
}

\affiliation[b]{
organization={School of Mathematics and Statistics},
addressline={Xi'an Jiaotong University},
city={Xi'an},
postcode={710049},
country={China}
}

\affiliation[c]{
organization={The University of Hong Kong},
addressline={Pokfulam Road},
city={Hong Kong},
postcode={999077},
country={China}
}

\address[d]{School of Computer Science, Duy Tan University, Da Nang 550000, Vietnam}

\address[e]{School of Computer Sciences, Universiti Sains Malaysia, Penang 11800, Malaysia}

\begin{abstract}
The development of multi-modal models has been rapidly advancing, with some demonstrating remarkable capabilities. However, annotating video-text pairs remains expensive and insufficient. Take video question answering (VideoQA) tasks as an example, human annotated questions and answers often cover only part of the video, since the corresponding text is often short and monotonous, leading to underutilization of video. To address this, we propose a \textbf{Bo}otstrapping \textbf{Vi}deo-\textbf{L}anguage \textbf{A}lignment framework (\textbf{BoViLA}), a self-training method that augments question samples during training process through LLM-based self-questioning and answering, which help model exploit video information and the internal knowledge of LLMs more thoroughly to improve modality alignment. However, low-quality self-generated questions may instead contaminate the performance, especially in the early stages of training, as we have observed in our experiments. To filter bad self-generated questions, we introduce Evidential Deep Learning (EDL) to estimate uncertainty and assess the quality of self-generated questions by evaluating the modality alignment within the context. To the best of our knowledge, this work is the first to explore LLM-based self-training frameworks for modality alignment. We evaluate BoViLA on five strong VideoQA benchmarks, where it outperforms several state-of-the-art methods and demonstrate its effectiveness and generality. Additionally, we provide extensive analyses of the self-training framework and the EDL-based uncertainty filtering mechanism. The code will be made available.
\end{abstract}


\begin{highlights}
\item Low data utilization and inefficient modality alignment in existing video-text methods motivate a novel self-training framework.
\item A self-questioning and answering training environment is constructed to enhance video-language alignment.
\item Evidential Deep Learning (EDL) is improved and successfully applied to LLMs for uncertainty estimation.
\item An EDL-based uncertainty-guided mechanism is devised to assess and filter self-generated questions.
\item The framework achieves higher data efficiency and improved modality alignment without additional human annotations.
\end{highlights}

\begin{keyword}


Self-questioning and answering \sep Large language models \sep Evidential deep learning \sep Modality alignment
\end{keyword}

\end{frontmatter}


\section{Introduction}

Recent advances in multimodal large models (MLLMs) have demonstrated the effectiveness of scaling laws in visual instruction fine-tuning. However, the continued advancement of MLLMs is hindered by the high cost of human annotation required for visual instruction data. In fact, this supervised fine-tuning paradigm does not fully exploit the rich information available in the visual modality data or the internal knowledge of frozen large language models (LLMs), yet merely increasing the training data without optimizing data utilization can be inefficient. For example, in conventional video question-answering (VideoQA) tasks, models typically predict answers based on a given video and its associated annotated question. This approach, however, is suboptimal for effective learning. On one hand, as noted in \citep{video-million-words} that “A Video Is Worth 1.8 Million Words”, videos often contain extensive information that can be described in various forms of language. However, typical datasets offer text that is both limited in length and uniform in structure, which significantly underutilizes the rich information embedded in videos. This restricts the model’s learning to the specific annotated question-answer pairs, limiting its ability to attend to the whole video and generalize to semantically similar questions. Consequently, this naive training paradigm hinders the model's learning efficiency.
On the other hand, this mechanical and passive supervised training method is notably inferior compared to human learning processes, which tends to be more active. Humans often draw upon past experiences and cognition to enrich their understanding of current events, hence proactively pose new questions and seeking answers for a more comprehensive and profound grasp of the situation, bringing higher learning efficiency by knowledge reusing.

\begin{figure*}[htbp]
\centering
\includegraphics[width=.85\columnwidth]{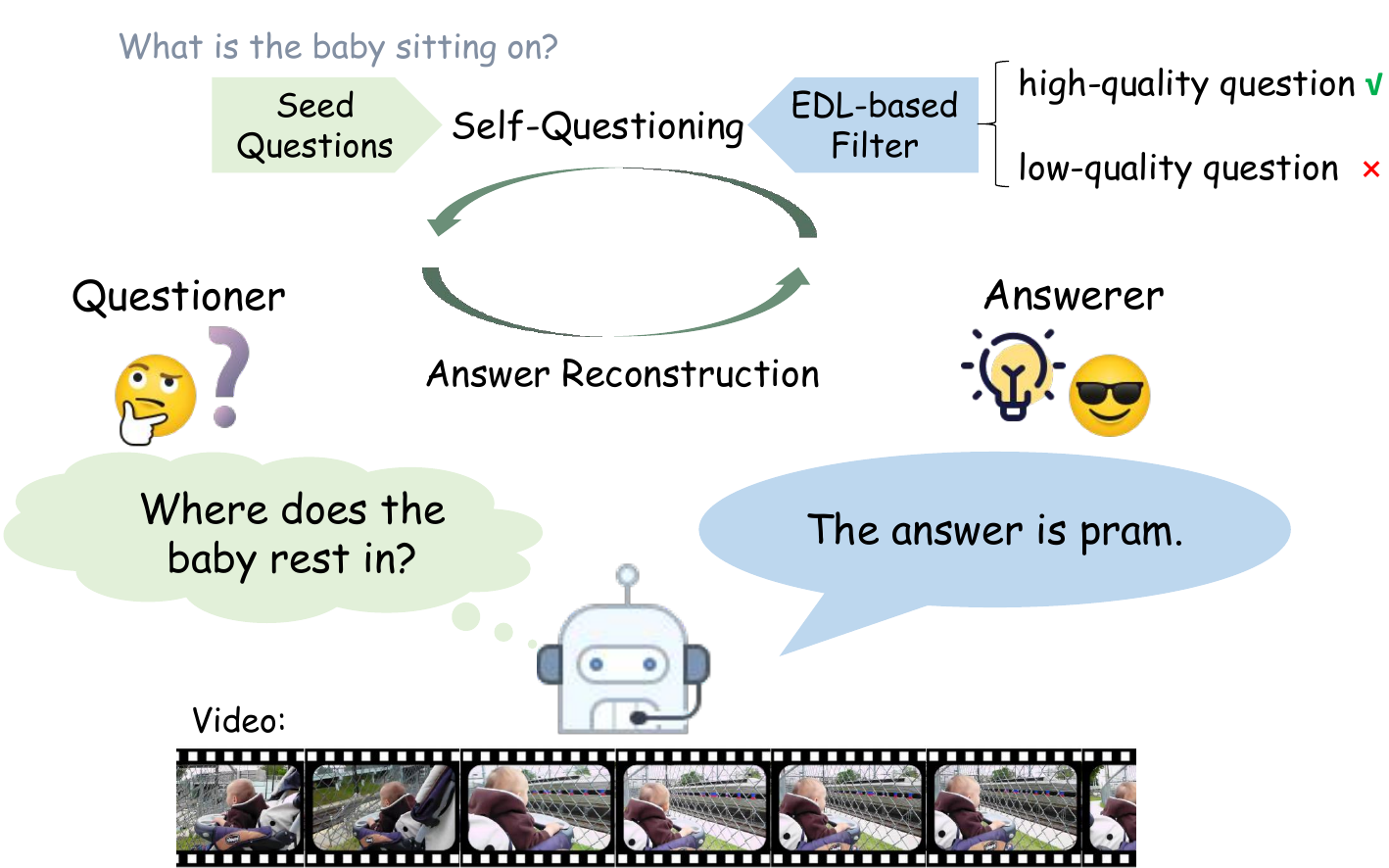}
\caption{\textbf{Framework overview.} The model plays the roles of both questioner and answerer. As a questioner, the model generates new questions based on the video, answer and seed question. As an answerer, the model endeavors to predict the answer from its own generated questions based on the video. Low-quality self-generated questions are filtered by an EDL-based filter to ensure that the knowledge received by the answerer is correct.}
\label{fig1}
\end{figure*}

To address this, we introduce \textbf{Bo}otstrapping \textbf{Vi}deo-\textbf{L}anguage \textbf{A}lignment (\textbf{BoViLA}) training framework via LLM-based self-questioning and answering. It further exploits rich information in the videos by reusing internal knowledge of LLM itself. BoViLA includes two roles, the \textbf{questioner} and the \textbf{answerer}, both played by the same model and improve each other alternately through self-questioning and answering, as shown in Fig.~\ref{fig1}. Questioner generates new questions for enabling itself to further extract aligned knowledge from videos and unleash power of the LLMs. Answerer provides feedback to questioner in terms of the self-generated questions. This framework features efficiently employment of video data and the internal historical knowledge of LLMs.

However, especially in the early stages of training, inadequate modal alignment often causes the questioner to generate low-quality questions. This makes training process unstable and may damage the final training results. To address this issue, we introduce an uncertainty estimation method based on evidential deep learning (EDL) to evaluate the uncertainty of LLMs' generation. Then, we control the influence our self-generated questions based on measured uncertainty. To effectively apply EDL to LLMs, whose most parameters are frozen and logits have high dimensions, we employ a technique called ``evidence decoupling'' to prevent information loss and ensure the effectiveness of uncertainty estimation. 

We verify the effectiveness of the BoViLA on five challenging VideoQA benchmarks: STAR \citep{star}, How2QA \citep{how2qa}, DramaQA \citep{dramaqa}, TVQA \citep{tvqa}, and VLEP \citep{vlep}, where BoViLA outperforms several strong baselines. Moreover, we present extensive ablation studies
as shown in Table~\ref{tab:ablation}. To sum up, our contributions are as follows:
\begin{itemize}
\item We propose a bootstrapping video-language alignment framework BoViLA, which help effectively enhance modality alignment via self-questioning and answering.
\item  We first investigate the uncertainty quantification method for LLMs based on Evidential Deep Learning (EDL), improving the vanilla EDL for LLMs by decoupling the direction and magnitude of evidence vector.
\item We validate the efficacy of BoViLA on five VideoQA benchmarks by outperforming several strong baseline models with only a few trainable parameters (4.5M). We also conduct thorough and detailed experiments to demonstrate the effectiveness of each component within BoViLA.
\end{itemize}

\section{Related Work}

\subsection{LLMs for multi-modal understanding}
As LLMs have demonstrated impressive capabilities \citep{gpt3, instructgpt, llama}, there has been increasing interest in exploring multi-modal language models (MLLMs) with visual capabilities. Unlike the more costly joint visual-language pre-training methods, some approaches focus on training lightweight visual-language connectors that endow LLMs with visual abilities \citep{ma2023llavilo}. These methods efficiently utilize the linguistic knowledge of LLMs and the visual knowledge of pre-trained visual encoders for cross-modal alignment. 
For instance, Flamingo \citep{flamingo} utilize a cross-attention mechanism to inject visual knowledge into the LLM, which is the so-called Perceiver Resampler. LLaMA-Adapter \citep{llama-adaptor} applies a linear projection along with prompt adaptation to incorporate visual information, effectively projecting visual embeddings into the input space of the LLM. Additionally, BLIP-2 \citep{blip2} trains a module called the Q-former to bridge the modal gap between pre-trained visual encoders and LLMs, enhancing the model's multi-modal understanding.

\subsection{Video Question-Answering}
VideoQA involves answering natural language questions about a video, requiring models to understand both the video and the questions across various semantic levels due to the open-ended nature of the questions, and answer them with commonsense reasoning. This makes VideoQA one of the most typical tasks in multi-modal understanding.

Traditional VideoQA methods relied on training separate visual and text encoders, along with temporal modeling and answering modules \citep{qian2023locate, xiao2022video, lei2021less}. However, with the advent of large language models (LLMs), there is a growing trend towards using LLM-based approaches due to their advanced reasoning abilities \citep{sevila, vlap, flipvqa, crema}. For instance, SeViLA \citep{sevila} uses an LLM to select keyframes from a video and employs another LLM to answer questions based on these keyframes, fully leveraging LLMs' capabilities. VLAP \citep{vlap} improves on this by introducing a Frame-Prompter and QFormer-Distiller for more efficient modality alignment. CREMA \citep{crema} trains a multimodal Q-former to integrate information from different modalities to answer questions. We notice the recent work of LLaMA-VQA \citep{flipvqa} as closest to ours, which trains only a linear layer and adaptor to enable LLMs to understand videos, and enhancing modality alignment through multi-task learning by reconstructing both questions and video. In contrast, our work prompts LLMs to simultaneously engage in self-questioning and answering. The major differences are that \textbf{(i)} We use self-generated questions as augmented training data and ask the model to answer them correctly. \textbf{(ii)} We further enhance the model's questioning ability by encouraging it to answer correctly from self-generated questions. In other words, the loss of the answerer on the self-generated questions propagates gradients back to the parameters of the questioner.

\subsection{LLM-Based Bootstrapping Training}
Early research has extensively explored the use of LLMs to generate training data for other models, which leverages the capabilities of LLMs to reduce the dependency on human labor for data collection \citep{lee2024llm2llm, huang2022large, wang2024vigc}.
Recently, there has been increasing interest in using LLMs to generate data for their own training \citep{ulmer2024bootstrapping, zhao2024self, self-instruct}. 
For example, \citep{self-instruct} enables models to generate data for instruction fine-tuning, focusing on data efficiency and general-purpose tasks. \citep{huang2022large} use Chain-of-Thought prompting and self-consistency to generate rationale-augmented answers without labeled data. \citep{ulmer2024bootstrapping} focuses on a single improve step and employs a conceptually simpler supervised finetuning strategy instead of RL. \citep{zhao2024self} investigates how self-generation can further enhance an instruction-finetuned model's ability to execute task-specific instructions. Our work differs from prior efforts in several key ways: \textbf{(i)} We focus on multi-modal tasks rather than text-only ones; \textbf{(ii)} We update model parameters end-to-end rather than alternately performing these processes offline to improve modality alignment; \textbf{(iii)} This dual approach enhances learning efficiency, offering a faster and more convenient solution.

Furthermore, training data generation can be seen as a form of knowledge distillation \citep{self-instruct, yang2024self}, while our self-questioning and answering method can also be seen as self-distillation.

\subsection{Uncertainty Estimation Model}
Numerous studies have explored models capable of estimating uncertainty to enhance reliability and trustworthiness \citep{amini2020deep, sensoy2018evidential}. EDL \citep{sensoy2018evidential}, as one of these approaches, models "second-order probabilities" over logits based on Dempster-Shafer Theory \citep{shafer1992dempster} and Subjective Logic \citep{jsang2018subjective} to capture uncertainty conveniently and accurately in various fields \citep{shao2024dual, holmquist2023evidential, huang2024crest}, particularly for out-of-distribution (OOD) samples. In this work, we leverage EDL-estimated uncertainty to evaluate the modality alignment within the context and employ it for "soft filtering" of the model's self-generated questions. To the best of our knowledge, we are the first to explore the integration of EDL with LLMs.

\section{Methdology}
We first present the architecture of BoViLA, detailing its key components and functionalities. Then we elaborate on our novel bootstrapping training framework, which leverages self-questioning and answering to enhance learning efficacy and modality alignment. Finally, we outlines our innovative approach for filtering self-generated questions, which is based on EDL-estimated uncertainty.

\subsection{Model Architecture}

\begin{figure*}[tbpt]
\centering
\includegraphics[width=.85\textwidth]{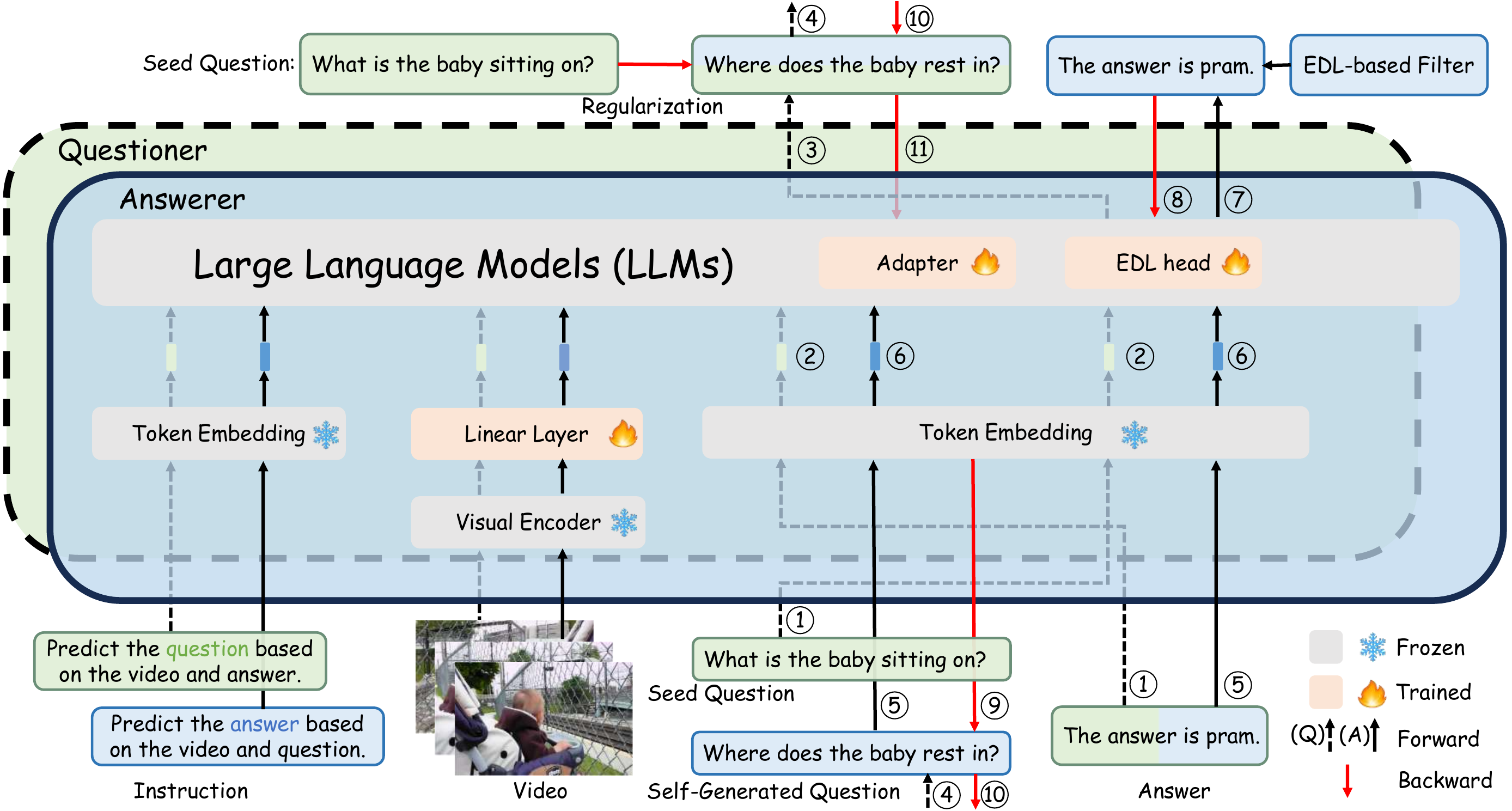}
\caption{\textbf{Model overview.} Our model acts as both questioner and answerer. During the forward pass, the questioner generates new questions from the seed question, which are then used as input for the answerer. Green elements and dashed arrows are associated with questioner, while blue elements and solid arrows pertain to answerer. In the backward pass, the answerer backpropagates gradients from the self-generated questions to the questioner, as shown by the red arrows. The self-generated questions are constrained by regularization and EDL-based filter. Steps 1-11 illustrate the BoViLA workflow, detailing the question-answer bootstrapping process.}
\label{fig2}
\end{figure*}

As shown in Fig.~\ref{fig2}, our model architecture consists of an LLM decoder, a learnable linear layer for mapping visual tokens to the text embedding space, a lightweight adaptor for task-specific fine-tuning, and an EDL head to estimate uncertainty. For videos, we first extract frames $\bm{v}=\{v_1, v_2, \cdots, v_{N_v}\}$ and use a pretrained visual encoder $E(\cdot)$ to extract their features $E(\bm{v})=\{E(v_1), E(v_2), \cdots, E(v_{N_v})\}\in{\mathbb{R}}^{{N_v}\times D}$. These features are then mapped to the text embedding space with the learnable linear layer $f_{\theta}(\cdot)$, and an extra learnable temporal embedding $\bm{t}=\{t_1, t_2, \cdots, t_{N_v}\}\in{\mathbb{R}}^{{N_v}\times D}$ is added, \emph{i.e.} 
\begin{align}
h_v&=f(E(\bm{v}))+\bm{t}
\\&=\{f(E(v_1))+t_1, 
\cdots, f(E(v_{N_v}))+t_{N_v}\}
\in{\mathbb{R}}^{{N_v}\times D}. 
\end{align} 
For text inputs such as task instructions, questions, or answers, we use the LLM's tokenizer and token embedding module to obtain the corresponding tokens and embedding. Specifically, for questions, we get the tokens $\bm{q}=\{q_1, q_2, \cdots, q_{N_q}\}$ and their embedding $\bm{h}^0_q=\{q^0_1, q^0_2, \cdots, q^0_{N_q}\}$. Similarly, for answers, we obtain the tokens $\bm{a}=\{a_1, a_2, \cdots, a_{N_a}\}$ and their embedding $\bm{h}^0_a=\{a^0_1, a^0_2, \cdots, a^0_{N_a}\}$.
Then, we concatenate the video and text tokens together as input to the LLM. As task-specific fine-tuning is necessary, we employed several prevalent PEFT methods such as LoRA \citep{lora}, adapter methods \citep{adapters, llama-adapter}, and prefix-tuning \citep{prefix-tuning, llama-adapter} for efficient modal alignment.
\subsection{Bootstrapping Training Framework}
In our bootstrapping training framework, the model acts as both a "questioner" and an "answerer". The questioner generates additional questions samples for the answerer, and the answerer improves its answering skills by tackling with these questions while providing feedback about quality of questions to improve the questioner's ability. The overall framework is illustrated in Fig. 1.
\subsubsection{Questioner.}
When the model acts as the questioner, we instruct it to generate a question based on the video, answer, and seed question, as shown in Figure~\ref{fig2}. Assuming the LLM has $l$ layers, the hidden states from the final layer are $h^l_q=\{q^l_1, q^l_2, \cdots, q^l_{N_q}\}$. For gradient backpropagation, we use Gumbel-Softmax \citep{gumbel-softmax} for sampling, formulated as below:
\begin{align}
\label{questioner}
p(\bm{\overline{q}}|\bm{v}, \bm{a}, \bm{q}) &= \prod_{i=1}^{N_q} p(\overline{q_i}|\bm{v}, \bm{a}, q_{<i}) \\
&= \prod_{i=1}^{N_q} \text{Gumbel-Softmax}(\text{Linear}(\overline{q}^l_i)). 
\end{align}
As Gumbel-Softmax can output one-hot probability vector, we directly multiply it with the token embedding matrix to obtain a gradient-propagatable question embedding $h^0_{\overline{q}}=\{{\overline{q}}^0_1, {\overline{q}}^0_2, \cdots, {\overline{q}}^0_{N_q}\}$, which serves as the input for the answerer.

\subsubsection{Answerer.}
When the model acts as the answerer, we ask it to predict answers of all questions (seed questions and self-generated questions) based on the video, where the answers share those of seed questions because the questioner generates questions conditioned on these answers. To be detailed, the loss can be computed as:
\begin{equation}
\label{vqa}
\vspace{-2mm}
\mathcal{L}_{\mathrm{vqa}} = -\log p(\bm{a}|\bm{v},\bm{q}) = -\sum\nolimits_{i=1}^{N_{a}} \log p(a_{i}|\bm{v},\bm{q},a_{<i}). 
\end{equation}
\begin{equation}
\label{aqa}
\mathcal{L}_{\mathrm{v\overline{q}a}} = -\log p(\bm{a}|\bm{v},\bm{\overline{q}}) = -\sum\nolimits_{i=1}^{N_{a}} \log p(a_{i}|\bm{v},\bm{\overline{q}},a_{<i}). 
\end{equation}

We provide the detailed prompt template of BoViLA in Table~\ref{tab:prompt}.

\begin{table}[H]
\caption{\textbf{Input prompt of questioner and answerer}.}
\centering
\begin{tabular}{@{}p{0.45\linewidth}@{\hspace{0.04\linewidth}}p{0.45\linewidth}@{}}
\begin{tcolorbox}[colback=lightgreen!60,
                  colframe=gray,
                  width=\linewidth,
                  arc=1mm, auto outer arc,
                  title=Questioner Template:,
                  boxrule=0.5pt,
                 ]
                {\tt [SOS] Video:} $\langle v_1 \rangle$ $\langle v_2 \rangle$ $\cdots$ $\langle v_{N_v} \rangle$ \\
                {\tt Choices:} \\ 
                (A) $\langle$ option 1 $\rangle$ \\
                (B) $\langle$ option 2 $\rangle$ \\ 
                (C) $\langle$ option 3 $\rangle$ \\
                (D) $\langle$ option 4 $\rangle$ \\
                (E) $\langle$ option 5 $\rangle$ \\
                {\tt Answer:} The answer is $\langle$ answer $\rangle$ {\tt [EOS]} \\
                {\tt Question:} \textcolor{red}{$\langle$ self-generated question $\rangle$ {\tt [EOS]}}
\end{tcolorbox}
&
\begin{tcolorbox}[colback=cyan!10,
                  colframe=gray,
                  width=\linewidth,
                  arc=1mm, auto outer arc,
                  title=Answerer Template:,
                  boxrule=0.5pt,
                 ]
                {\tt [SOS] Video:} $\langle v_1 \rangle$ $\langle v_2 \rangle$ $\cdots$ $\langle v_{N_v} \rangle$ \\
                {\tt Question:} \textcolor{red}{$\langle$ self-generated question $\rangle$} \\
                {\tt Choices:} \\ 
                (A) $\langle$ option 1 $\rangle$ \\
                (B) $\langle$ option 2 $\rangle$ \\ 
                (C) $\langle$ option 3 $\rangle$ \\
                (D) $\langle$ option 4 $\rangle$ \\
                (E) $\langle$ option 5 $\rangle$ \\
                {\tt Answer:} The answer is \textcolor{red}{$\langle$ answer $\rangle$ {\tt [EOS]}}
\end{tcolorbox}
\end{tabular}
\label{tab:prompt}
\end{table}

\subsubsection{Regularization Based on Seed Questions.}
\label{reg}
A key point of BoViLA is the differentiability of self-generated questions. If answerer fails to predict the target answer from $\overline{q}$, then penalty will be applied to the questioner to improve itself. In contrast, if answerer succeed in answering $\overline{q}$, that means $\overline{q}$ is considered a "good" question by answerer. However, this can easily lead to "information leakage", where the questioner creates meaningless questions but contain target answers implicitly. To address this, we apply seed-based question regularization to constrain the generation space of questioner as follows:
\begin{align}
\label{reg}
\mathcal{L}_{\mathrm{reg}}
&=KL[p(\bm{q}) \mid\mid p(\bm{\overline{q}}|\bm{v}, \bm{a}, \bm{q})] \\
&=\sum_{i=1}^{N_{q}}KL[p(q_i) \mid\mid p(\overline{q}_i|\bm{v},\bm{a},q_{<i})] \\
&=\sum_{i=1}^{N_{q}} p(q_i) \log \frac{p(q_i)}{p(\overline{q}_i|\bm{v},\bm{a},q_{<i})}. 
\end{align}

Detailed ablation experiments are presented in Section~\ref{ablation}, qualitatively and quantitatively demonstrating the effectiveness of each design, including the trainging with augmented question-answer samples and the regularization trick.

\subsection{EDL-based Filter}
Despite the effectiveness of regularization, the questioner can still generate low-quality questions due to modal unalignment (especially at the early stages of training). Since EDL excels at predicting high uncertainty for OOD samples, which lie beyond the model’s knowledge, and we consider misaligned video contexts and low-quality questions can be regarded as OOD samples, we introduce EDL-estimated uncertainty to assess the degree of modality alignment and the quality of self-generated questions. We then adjust the impact on loss $\mathcal{L}_{\mathrm{v\overline{q}a}}$ based on the level of uncertainty.

Vallina EDL treats transformed logits as strength parameters $\bm{\alpha}=(\alpha_1, \alpha_2, \cdots, \alpha_{K})$ of a Dirichlet distribution $\mathrm{Dir}(\bm{p}|\bm{\alpha})$ in a $K$-way classification and samples class probabilities $\bm{p}=(p_1, p_2, \cdots, p_{K})$ from this distribution as the final prediction. It is also applicable to LLMs, as they generate text via next-token prediction, which can essentially be viewed as $K$-way classification where $K$ is great.

However, directly applying vanilla EDL to LLMs would fail. This is because vanilla EDL commonly applies non-negative activation functions like ReLU \citep{sensoy2018evidential} or Softplus \citep{amini2020deep} on logits to ensure the non-negativity of evidence. This causes information loss and negative effects \citep{ye2024uncertainty, meinert2023unreasonable, wu2024evidence}, especially for finetuning pretrained models with a large number of categories. To overcome this and successfully apply EDL to LLMs, we propose evidence decoupling to decouple the direction and magnitude of the evidence vector $\bm{e}=(e_1, e_2, \cdots, e_{K})$ to mitigate information loss. To be detailed, assume the logits output by the model are $\bm{z}=(z_1, z_2, \cdots, z_{K})$, vanilla EDL determines $\bm{\alpha}$ through the following transformation: 
\begin{equation}
\label{vanilla-edl}
\alpha_i = e_i + 1 = \text{ReLU}(z_i) + 1,\quad i = 1, 2, \cdots, K, 
\end{equation}
and the total strength $S$ can be evaluated as $S=\sum_{i=1}^{K}\alpha_{i}$. Dirichlet distribution is modeled with $\bm{\alpha}$ as:
\begin{equation}
\label{dir}
\mathrm{Dir}(\bm{p}|\bm{\alpha}) = \left\{\begin{array}{ll}\frac{1}{B(\bm{\alpha})}\prod_{i=1}^Kp_i^{\alpha_i-1}&\bm{p}\in\mathcal{S}_K,\\0&\mathrm{otherwise},\end{array}\right.
\end{equation}
where $\mathcal{S}_K$ is the $K$-dimensional unit simplex,
\begin{equation}
\label{simplex}
\mathcal{S}_K=\left\{\bm{p}\bigg|\sum_{i=1}^Kp_i=1, 0\leq p_1,\cdots,p_K\leq1\right\}, 
\end{equation}
and $B(\bm{\alpha})$ is the $K$-dimensional multinomial beta function.
We propose evaluating direction $\bm{d}=(d_1, d_2, \cdots, d_{K})$ and magnitude, \emph{i.e.} total strength $S$ of evidence respectively to mitigate information loss. As to direction, we apply softmax function to the logits for non-negativity, which preserves the internal distribution structure of the logits:
\begin{equation}
\label{direction}
d_i = \frac{e^{z_i}}{\sum_{j=1}^{K} e^{z_j}},\quad i = 1, 2, \cdots, K. 
\end{equation}
Since $\sum_{i=1}^{K} d_i = 1$, as is the so-called "direction", we have to evaluate magnitude in other ways. Due to the large number of categories, the target probability vector is inherently sparse, making the uniform calculation of $S$, as in vanilla EDL, suboptimal. We train a simple linear layer, named as EDL head, to capture this sparsity and use a sigmoid function along with simple mathematical transformation to obtain a non-negative magnitude without loss:
\begin{equation}
\label{magnitude}
S = \frac{\text{sigmoid}(\text{Linear}(z))}{1-\text{sigmoid}(\text{Linear}(z))} \in (0, \infty). 
\end{equation}
The final evidence and strength can be formulated as follows:
\begin{equation}
\label{llm-edl}
\alpha_i = e_i + 1 = S\cdot d_i + 1,\quad i = 1, 2, \cdots, K, 
\end{equation}
For training EDL head, we expand $\mathcal{L}_{\mathrm{vqa}}$ to form of expectation:
\begin{align}
\label{edl-loss}
\mathcal{L}_{\mathrm{vqa}}^{\mathrm{edl}}
&=E_\mathrm{Dir}[-\log p_{j{a_j}}] \\
&=-\sum_{j=1}^{N_{a}}\int\log p_{j{a_j}}\frac{1}{B(\boldsymbol{\alpha}_{j})}\prod_{i=1}^{K}p_{ji}^{\alpha_{ji}-1}d\bm{p}_{j}, 
\end{align}
where $$p_{ji}=p(\text{word}_i|\bm{v},\bm{q},a_{<j}),\quad p_{j{a_j}}=p(a_j|\bm{v},\bm{q},a_{<j}).$$We also apply the regularization loss $\mathcal{L}_{\mathrm{reg}}^{\mathrm{edl}}$ mentioned in \citep{sensoy2018evidential}. After training, EDL uncertainty can be estimated as:
\begin{equation}
\label{uncertainty}
u=\frac{\sum_{i=1}^{N_{a}}\frac{K}{S_i}}{N_{a}}, 
\end{equation}
where $S_i$ represents total strength corresponding to $p(\cdot|\bm{v},\bm{q},a_{<i})$. 

To the best of our knowledge, our work is the first to successfully apply EDL to LLMs, offering a more convenient method for measuring the uncertainty of LLMs. We conduct more detailed ablation studies and analysis on the effectiveness of our EDL-based filter and the evidence decoupling technique in Section~\ref{edl} and the supplementary materials. 

This uncertainty is then used to filter self-generated questions of low-quality by simply controling the weight of loss $\mathcal{L}_{\mathrm{v\overline{q}a}}$ with $1-u$. Finally, we train BoViLA with the following total loss:
\begin{equation}
\label{total-loss}
\mathcal{L}_\mathrm{BoViLA} = \mathcal{L}_{\mathrm{vqa}}^{\mathrm{edl}} + (1-u)\cdot\mathcal{L}_{\mathrm{v\overline{q}a}} + \mathcal{L}_{\mathrm{reg}} + \mathcal{L}_{\mathrm{reg}}^{\mathrm{edl}}.
\end{equation}

\section{Experiment}
In this section, we outline our experimental setup and demonstrate the superiority 
of our BoViLA framework on 5 challenging VideoQA benchmarks.
Furthermore, we conduct extensive ablation studies to show the effectiveness of each component in our framework, including the questioner, answerer, EDL-based filter, and regularization based on seed questions. 
We also perform in-depth quantitative and qualitative analyses on our self-generated questions.

\subsection{Experimental Setup}

\subsubsection{Implementation Details.}
We conduct all training with 8 × 80GB A800 GPUs for 10 epochs. 
For all the datasets, we use VIT-L/14 as the visual encoder to extract 10 frame features for each video and use LLaMA(7B) as our large language model. 
Regarding evaluation metrics, we use the accuracy of choosing the right answer and test on the validation split. 
More details are provided in Appendix~\ref{tab:implementation}.

\subsubsection{Baselines \& Benchmarks.}
We compare our framework with some state-of-the-art (SOTA) baselines, especially that are LLM-based such as BLIP-2 \citep{blip2}, SeViLA \citep{sevila} and LLaMA-VQA \citep{flipvqa}, on 5 challenging multi-choice VideoQA benchmarks: 
1) TVQA \citep{tvqa}, requireing answer questions based on video, dialogues and scenes, featuring 152,545 QA pairs from 21,793 video clips extracted from popular TV shows.
2) STAR \citep{star}, which is designed for spatio-temporal and relational reasoning, containing 22,670 QA pairs based on 12,672 video clips.
3) DramaQA \citep{dramaqa}, which is tailored for emotional and social reasoning, featuring 16,191 QA pairs derived from 23,239 video clips. 
4) VLEP \citep{vlep}, which focuses on predicting future events based on video and dialogues, consisting of 28,726 QA pairs from 10,000 video clips.
5) How2QA \citep{how2qa}, which is designed for instructional video comprehension, containing 46,467 QA pairs derived from 23,228 video clips.

\subsection{Main Results}

\begin{sidewaystable}[htbp]
\label{table1}
    \centering
    \caption{
    \textbf{Comparison on five challenging VideoQA benchmarks with both LLMs-based and non-LLMs-based baselines.}
    STAR contains four question types: \textbf{Int.}(interaction), \textbf{Seq.}(sequence), \textbf{Pre.}(prediction), and \textbf{Fea.}(feasibility).
    * denotes that we do not use the speech captions.  
    Total accuracy is highlighted in green. 
    The best results in each column are highlighted in bold, while the second-best results are underlined, to clearly indicate the model's performance rankings across different datasets.
    }
    \begin{adjustbox}{width=0.99\textwidth}
   \begin{tabular}{c|c|c|c c c c >{\columncolor{lightgreen}} c| >{\columncolor{lightgreen}}c|>{\columncolor{lightgreen}} c| >{\columncolor{lightgreen}}c| >{\columncolor{lightgreen}}c }
        \toprule
        \multirow{2}{*}{\textbf{Models}} & \multirow{2}{*}{\textbf{Language Model}} & \multirow{2}{*}{\shortstack{\textbf{\# trainable} \\ \textbf{params}}}  & \multicolumn{5}{c|}{\textbf{STAR}} & \multicolumn{1}{c|}{\textbf{DramaQA}} & \multicolumn{1}{c|}{\textbf{VLEP}} & \multicolumn{1}{c|}{\textbf{TVQA*}} & \multicolumn{1}{c}{\textbf{How2QA}} \\ 
        & &  & Int. & Seq. & Pre. & Fea. & Tot. & Tot. & Tot. & Tot. & Tot. \\
        \midrule
        \midrule
    
        
        \multicolumn{1}{l|}{FrozenBiLM~\cite{yang2022zero}} & DeBERTa & 30M  & - & - & - & - & - & - & - & 57.5 & 86.7\\
        
        \multicolumn{1}{l|}{MERLOT~\cite{zha2019spatiotemporal}} & RoBERTa & 223M  & - & - & - & - & - & 81.4 & 68.4 & - & - \\
        
        
        \multicolumn{1}{l|}{SPCRL~\cite{kim2021self}} & BERT  & - & - & - & - & - & - & 81.0 & - & - & - \\
        
        
        \multicolumn{1}{l|}{AIO~\cite{wang2023all}} & - & 110M   & 47.5 & 50.8 & 47.8 & 44.1 & 47.5 & - & - & - & -\\
        
        \multicolumn{1}{l|}{ATP~\cite{buch2022revisiting}} & CLIP  & -  & 50.6 & 52.9 & 49.4 & 40.6 & 48.4 & - & - & - & - \\
        
        \multicolumn{1}{l|}{MIST~\cite{gao2023mist}} & -  & - & 55.6 & 54.2 & 54.2 & 44.5 & 53.9 & - & - & - & -\\
        
        
        \multicolumn{1}{l|}{InternVideo~\cite{wang2022internvideo}} & CLIP  & 1.3B  & 62.7 & 65.6 & 54.9 & 51.9 & 58.7 & - & 63.9 & 57.2 & 79.0 \\
        
        \multicolumn{1}{l|}{LLaMa-VQA~\cite{flipvqa}} & LLaMA & 4.5M  & 66.2 & 67.9 & 57.2 & 52.7 & 65.4 & \underline{84.1} & \underline{71.0} & \underline{70.4} & - \\

        \multicolumn{1}{l|}{BLIP-2~\cite{blip2}} &  Flan-T5 & 432M & 52.3 & 54.8 & 49.0 & 51.2 & 51.8 & - & 67.0 & 54.5 & 82.2 \\

        \multicolumn{1}{l|}{SeViLA~\cite{sevila}} &  Flan-T5 & 216M & 63.7 & \textbf{70.4} & \underline{63.1} & \textbf{62.4} & 64.9 & - & 68.9 & 61.6 & 83.6 \\
        
        \multicolumn{1}{l|}{VLAP~\cite{vlap}} &  Flan-T5 & 188M  & \textbf{70.0} & \textbf{70.4} & \textbf{65.9} & \underline{62.2} & \textbf{67.1} & -  & 69.6 & 63.4 & \underline{83.9} \\
        \midrule
        
        \multicolumn{1}{l|}{\textbf{BoViLA} (Ours)} & LLaMA & \textbf{4.5M}  & \underline{66.9} & \underline{68.0} & 62.0 & 57.2 & \underline{66.4} & \textbf{85.2} & \textbf{71.2} &  \textbf{71.6} & \textbf{89.4} \\
        
        
        \bottomrule
    \end{tabular}
    \end{adjustbox}
    \label{tab:main}
\end{sidewaystable}
Table~\ref{tab:main} shows comparison results between our BoViLA and several strong baseline methods on the VideoQA task. Our proposed BoViLA achieves superior performance across multiple VideoQA benchmark datasets, showcasing strong capabilities in 1) cross-modal understanding of descriptive questions and 2) advanced temporal causal reasoning. To begin with, take How2QA benchmark as an example, which focuses on understanding video content and tests the model's ability to perceive detailed visual information based on the given questions. We consider such ability as a fundamental capability for video-textual cross-modal understanding. Our model outperforms the state-of-the-art by more than 2.7\%. For the more demanding task of temporal causal reasoning, we report the results on the STAR, DramaQA, VLEP, and TVQA datasets in Table~\ref{tab:main}. For example, our method outperforms the previous best model by 8.2\% in total accuracy on the TVQA dataset and exceeds the performance of existing models by 1.1\% on the DramaQA dataset.

It is noteworthy that, compared to many VideoQA baselines utilizing large language models (LLMs), our approach enhances reasoning capabilities on VideoQA benchmarks with only 4.5M trainable parameters. This efficiency is achieved through our Bootstrapping training method, which effectively leverages the strong priors provided by LLMs. As a result, BoViLA not only significantly reduces training costs compared to models trained from scratch (e.g., InternVideo, 1.3B), but also outperforms most parameter-efficient fine-tuning (PEFT) paradigms.
\subsection{Ablation Studies}
\begin{table}[htbp]
\centering
\caption{\textbf{Ablation studies about BoViLA framework on STAR validation dataset}. The \textbf{vqa}, \textbf{reg}, \textbf{v$\overline{\text{q}}$a} and \textbf{EDL} respectively represents 
$\mathcal{L}_{\mathrm{vqa}}$, 
$\mathcal{L}_{\mathrm{reg}}$, 
$\mathcal{L}_{\mathrm{v\overline{q}a}}$, 
and EDL-based filter. 
"GP" means the answerer can backpropagate the gradient to the questioner, while the "NGP" refers to the opposite.}
\begin{tabular}{c c c c | c c c c c}
\toprule
\multirow{2}{*}{\textbf{vqa}} & \multirow{2}{*}{\textbf{reg}} & \multirow{2}{*}{\textbf{v$\overline{\text{q}}$a}} & \multirow{2}{*}{\textbf{EDL}} & \multicolumn{5}{c}{\textbf{STAR}}\\
\cmidrule{5-9}
 &  &  &  & Int. & Seq. & Pre. & Fea. & Avg. \\
\midrule
\checkmark & - & - & - & 65.6 & 66.0 & 54.7 & 54.9 & 64.1 \\
\checkmark & \checkmark & - & - & 66.0 & 66.8 & 59.1 & 54.9 & 65.0 \\
\checkmark & \checkmark & \multicolumn{1}{l}{\checkmark(NGP)} & - & 66.6 & 67.5 & 58.7 & \underline{56.7} & 65.7 \\
\checkmark & \checkmark & \multicolumn{1}{l}{\checkmark(NGP)} & - & \textbf{67.0} & \underline{67.7} & \underline{60.0} & 54.8 & \underline{65.9} \\
\checkmark & \checkmark & \multicolumn{1}{l}{\checkmark(GP)} & \checkmark & \underline{66.9} & \textbf{68.0} & \textbf{62.0} & \textbf{57.2} & \textbf{66.4} \\
\bottomrule
\end{tabular}
\label{tab:ablation}
\end{table}

\begin{figure}[htbp]
\centering
\includegraphics[width=.9\columnwidth]{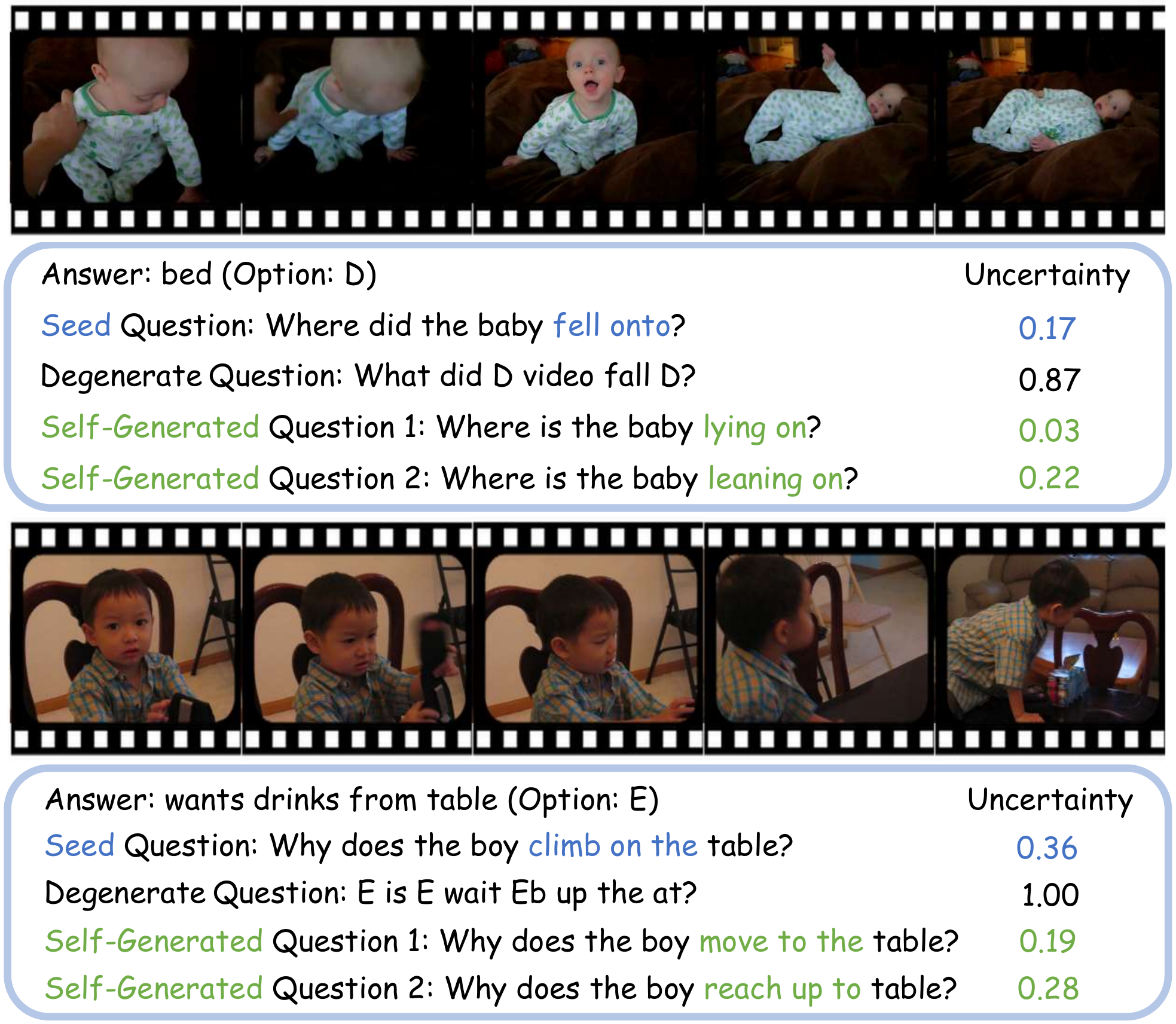}
\caption{\textbf{Examples of vanilla self-generated questions (degenerate question), self-generated questions with regularization term, and the corresponding EDL-estimated uncertainty.}}
\label{fig3}
\end{figure}

\label{ablation}
The self-questioning and answering process can easily cause "information leakage", where the questioner creates degenerate questions, which are meaningless and cheatingly contain target answers. Our regularization method and EDL-based filter really help solve this problem and allow the model to generate high-quality questions and filter low-quality questions to boost answers.
In Figure~\ref{fig3}, we qualitatively demonstrate the self-generated questions, the regularization term, and the EDL-estimated uncertainty through several examples. 
Seed Question refers to the questions labeled in the dataset, and Answer represents corresponding answers along with respective options. Degenerate Question refers to the junk questions generated by the model when regularization is not used. Self-Generated Question 1 and 2 represent high-quality questions generated by the model equipped with regularization. These questions are semantically consistent with the video and leverage the internal knowledge of the LLMs to provide diverse questions. The uncertainty is estimated by the EDL head. In these two examples, it is evident that the question quality is negatively correlated with the uncertainty. Additionally, it is worth mentioning that since we used the gumbel-softmax function with some randomness in sampling self-generated questions, different questions are generated for the same sample during each training epoch.

Furthermore, we provide comprehensive ablation studies on STAR validation set in Table~\ref{tab:ablation} to verify the effectiveness of our method quantitatively. The results clearly demonstrate: 1) Only learning from regularization about questioning allows the model to achieve a 2.2\% performance boost. 2) Further learning by answering these self-generated questions contributes an additional 0.7\% improvement. 3) If the answerer is allowed to backpropagate gradients to the questioner, \emph{i.e.} providing feedback on question quality, performance can increase by another 0.5\%. 4) Moreover, applying an EDL-based filter to progressively eliminate potential junk questions that could negatively impact the model can further enhance performance by 0.9\%.

\begin{table}[htbp]
\centering
\caption{\textbf{Ablation studies about our improved EDL method on STAR validation dataset}. The \textbf{Linear} and \textbf{Softmax} refer to the methods we introduce to evaluate total strength and evidence direction.}
\begin{tabular}{c c | c c c c c}
\toprule

\multirow{2}{*}{\textbf{Linear}} & \multirow{2}{*}{\textbf{Softmax}} & \multicolumn{5}{c}{\textbf{STAR}}\\
\cmidrule{3-7}
 & & Int. & Seq. & Pre. & Fea. & Avg. \\
\midrule
- & - & - & - & - & - & - \\
\checkmark & - & 24.7 & 23.1 & 22.9 & 21.6 & 23.5 \\
\checkmark & \checkmark & 66.9 & 68.0 & 62.0 & 57.2 & 66.4 \\
\bottomrule
\end{tabular}
\label{tab:edl}
\end{table}
We also conduct an ablation study on our improved EDL method. As shown in Table~\ref{tab:edl}, without our proposed linear projection to calculate the total strength (for evidence decoupling) yet simply sum it up, the model's training will break down. This is due to the large number of classes in the vocabulary (even though most words receive very small logits), and summing directly without considering sparsity can easily lead to numerical instability. Moreover, if we do not use our proposed Softmax-based method to decouple the computation of evidence direction, the significant information loss will also cause the model to fail to converge.

\subsection{More Discussion}
\label{edl}
\begin{figure}[h]
\centering
\includegraphics[width=0.9\columnwidth]{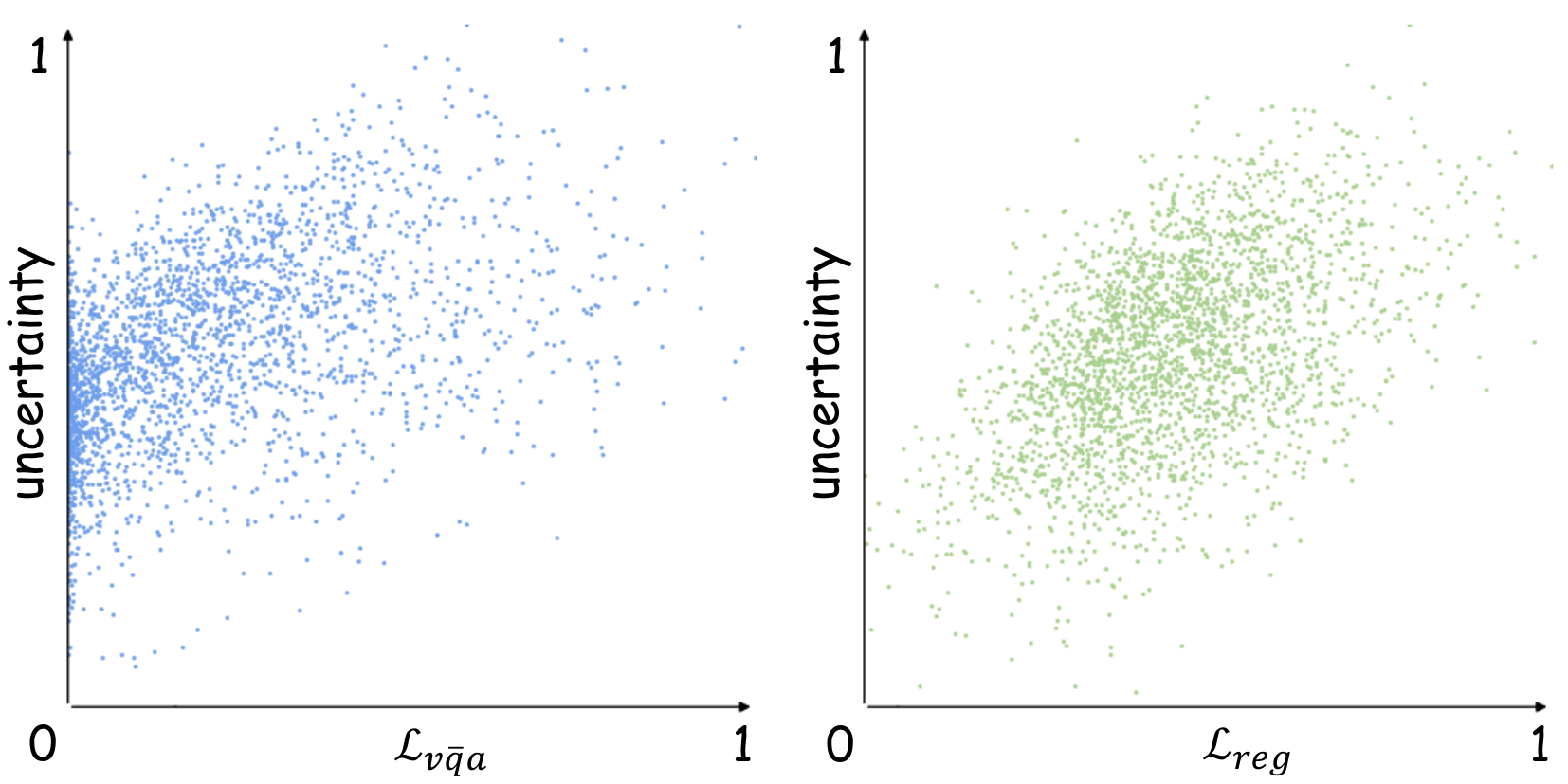}
\caption{\textbf{Correlation between EDL-estimated uncertainty and the quality of self-generated questions.} We use $\mathcal{L}_{\mathrm{v\overline{q}a}}$ and $\mathcal{L}_{\mathrm{reg}}$ to approximately represent the quality of self-generated questions. To conduct a clearer correlation analysis, we individually apply the Min-Max normalization to the uncertainty, $\mathcal{L}_{\mathrm{v\overline{q}a}}$ and $\mathcal{L}_{\mathrm{reg}}$, scaling them to the range of 0-1.}
\label{fig4}
\end{figure}

Our proposed EDL-based filter is built on the assumption that \textbf{“EDL-estimated uncertainty is able to measure the quality of self-generated questions and, is approximately negatively correlated with it"}. Here, we further explore and validate this assumption. We use $\mathcal{L}_{\mathrm{v\overline{q}a}}$ and $\mathcal{L}_{\mathrm{reg}}$ as approximate measures of the quality score for self-generated questions. $\mathcal{L}_{\mathrm{v\overline{q}a}}$ represents the similarity between the self-generated question $\overline{q}$ and the seed question $q$ in terms of KL divergence, where a high $\mathcal{L}_{\mathrm{v\overline{q}a}}$ indicates a completely uncontrolled self-generated question. $\mathcal{L}_{\mathrm{reg}}$ represents the likelihood that the answerer successfully predicts the target answer from the self-generated question, where a high $\mathcal{L}_{\mathrm{reg}}$ is likely to indicates that there is no meaningful semantics in the self-generated question. $\mathcal{L}_{\mathrm{v\overline{q}a}}$ and $\mathcal{L}_{\mathrm{reg}}$ complement and reinforce each other, so we use them together to represent the quality of self-generated questions. As shown in Figure~\ref{fig4}, both $\mathcal{L}_{\mathrm{v\overline{q}a}}$ and $\mathcal{L}_{\mathrm{reg}}$ demonstrate the pattern that \textbf{"the lower the quality of the self-generated question, the higher the uncertainty"}. This insight further provides a solid explanation for the effectiveness of our proposed EDL-based filter. Moreover, we also provide a more comprehensive analysis of the uncertainty distribution modeled by our improved EDL in Appendix~\ref{validity}, further supporting the effectiveness of the EDL-based filter.
\vspace{5mm}
\vspace{-8mm}

\section{Conclusion, Limitation and Future Work}
In this paper, we introduce BoViLA, a novel video-language alignment framework that leverages self-questioning and answering to autonomously generate new training samples, thereby enhancing data utilization. Notably, this process does not rely on additional annotated data, instead fully utilizing the rich information within videos and the internal knowledge of LLMs.

A key contribution of our framework is the improvement and integration of EDL method. EDL plays a crucial role in assessing and filtering the quality of self-generated questions, ensuring that the model focuses on high-quality, informative questions. By incorporating EDL, BoViLA improves the overall alignment between video and language modalities. This integration of EDL significantly boosts the framework's performance, allowing it to achieve superior results across five VideoQA benchmarks, outperforming current state-of-the-art methods with only a few trainable parameters.

Despite its achievements, BoViLA faces limitations due to its constrained question generation space. The questioner generates questions autoregressively based on the context of the seed question and answer, which restricts the diversity of the generated questions. Future work will focus on exploring methods to sample more diverse questions, aiming to further enhance the framework's capabilities and expand its application scope.

\appendix
\appendix

\onecolumn 
\renewcommand{\thetable}{\Alph{table}}
\renewcommand{\thefigure}{\Alph{figure}}
\renewcommand{\thesection}{\Alph{section}}
\renewcommand{\theequation}{\Alph{equation}}
\renewcommand{\thealgorithm}{\Alph{algorithm}}
\renewcommand{\thesubsection}{\thesection.\alph{subsection}}
\setcounter{figure}{0}    
\setcounter{table}{0}    
\setcounter{equation}{0}
\setcounter{algorithm}{0}
\setcounter{footnote}{0} 
\setcounter{page}{1}
\section*{Appendix}

\section{Implementation Details}
As reported in Table~\ref{tab:implementation}, we provide a detailed list of experimental settings across various datasets.

\begin{table}[htbp]
    \centering
    \caption{Summary of the datasets and implementation details used in the experiments, including dataset size, model settings, and training hyperparameters. \textbf{BS} denotes batch size. \textbf{LR} represents learning rate. \textbf{Max length} denotes the maximum number of tokens in the prompt. }
    \scalebox{0.8}{ 
        \begin{tabular}{l|c|cccccccc}
        \toprule
        \textbf{Dataset}  & \textbf{\string# Samples}  & \textbf{BS} & \textbf{LR} & \textbf{Epochs} & \textbf{Warm-up} & $\mathcal{L}_{\mathrm{v\overline{q}a}}$ & $\mathcal{L}_{\mathrm{reg}}$ & $\mathcal{L}_{\mathrm{reg}}^{\mathrm{edl}}$ & \textbf{Max length} \\
        \hline
        TVQA  & 122K & $4*8$ & $9e^{-2}$ & $5$ & $2$ & $0.05$ & $0.1$ & $1e^{-9}$ & $160$ \\
        STAR  & 45.7K &$4*8$ & $9e^{-2}$ & $10$ & $2$ & $0.25$ & $0.5$ & $1e^{-9}$ & $160$ \\
        DramaQA & 18.5K & $4*8$ & $9e^{-2}$ & $10$ & $2$ & $0.15$ & $0.3$ & $1e^{-9}$ & $256$ \\
        VLEP  & 20K & $4*8$ & $9e^{-2}$ & $10$ & $2$ & $0.25$ & $0.5$ & $1e^{-9}$ & $256$ \\
        How2QA & 34.2K & $4*8$ & $9e^{-2}$ & $3$ & $2$ & $0.3$ & $0.6$ & $1e^{-9}$ & $160$ \\
        \bottomrule
        \end{tabular}
    }

    \label{tab:implementation}
\end{table}


\section{Details of EDL Loss}
Here we show the detailed derivations of $\mathcal{L}_{\mathrm{vqa}}^{\mathrm{edl}}$ and $\mathcal{L}_{\mathrm{reg}}^{\mathrm{edl}}$. The $\mathcal{L}_{\mathrm{vqa}}^{\mathrm{edl}}$, which is essentially the Bayes risk, is as follows: 
\begin{align}
\mathcal{L}_{\mathrm{vqa}}^{\mathrm{edl}}
=\sum_{j=1}^{N_{a}}E_\mathrm{Dir}[-\log p_{j{a_j}}]
=\sum_{j=1}^{N_{a}}\int-\log p_{j{a_j}}\frac{1}{B(\boldsymbol{\alpha}_{j})}\prod_{i=1}^{K}p_{ji}^{\alpha_{ji}-1}d\bm{p}_{j}. 
\end{align}
By the properties of the expectation of the Dirichlet distribution, we have: 
\begin{align}
E_\mathrm{Dir}[\log p_{j{a_j}}] = \psi(\alpha_{j{a_j}}) - \psi(S_j),
\end{align}
where $\psi(\cdot)$ is the \textit{digamma} function. So the origin loss can be formulated as:
\begin{align}
\mathcal{L}_{\mathrm{vqa}}^{\mathrm{edl}}
=\sum_{j=1}^{N_{a}}-E_\mathrm{Dir}[\log p_{j{a_j}}]
=\sum_{j=1}^{N_{a}}\Big(\psi(S_{j})-\psi(\alpha_{j{a_j}})\Big).
\end{align}
The $\mathcal{L}_{\mathrm{reg}}^{\mathrm{edl}}$, which is essentially the KL divergence with the zero evidence Dirichlet distribution, is as follows: 
\begin{align}
\mathcal{L}_{\mathrm{reg}}^{\mathrm{edl}}
&=
KL[Dir(\boldsymbol{p}|\boldsymbol{\alpha})\|Dir(\boldsymbol{p}|\langle1,\ldots,1\rangle)]\\
&= E_{Dir(\boldsymbol{p}|\boldsymbol{\alpha})}\left[\log\frac{Dir(\boldsymbol{p}|\boldsymbol{\alpha})}{Dir(\boldsymbol{p}|\langle1,\ldots,1\rangle))}\right]\\
&=E_{Dir(\boldsymbol{p}|\boldsymbol{\alpha})}\left[\log Dir(\boldsymbol{p}|\boldsymbol{\alpha})-\log Dir(\boldsymbol{p}|\langle1,\ldots,1\rangle)\right)]\\
&=E_{Dir(\boldsymbol{p}|\boldsymbol{\alpha})}\left[-\log B(\boldsymbol{\alpha})+\sum_{k=1}^{K}(\alpha_{k}-1)\log p_k\right]
+
\log B(\langle1,\ldots,1\rangle)\\
&=\log\frac{B(\langle1,\ldots,1\rangle)}{B(\boldsymbol{\alpha})}+E_{Dir(\boldsymbol{p}|\boldsymbol{\alpha})}\left[\sum_{k=1}^{K}(\alpha_{k}-1)\log p_k\right]\\
&=\log\left(\frac{\Gamma(\sum_{k=1}^{K}\alpha_{k})}{\Gamma(K)\prod_{k=1}^{K}\Gamma(\alpha_{k})}\right)+\sum_{k=1}^{K}(\alpha_{k}-1)\bigg[\psi(\alpha_{k})-\psi\bigg(\sum_{j=1}^{K}\alpha_{k}\bigg)\bigg], 
\end{align}
where $\Gamma(\cdot)$ is the gamma function.

\section{Validity of EDL-Estimated Uncertainty}
\label{validity}
\subsection{Visualization of Uncertainty Distribution.}
\begin{figure}[t!]
\centering
\includegraphics[width=0.5\columnwidth]{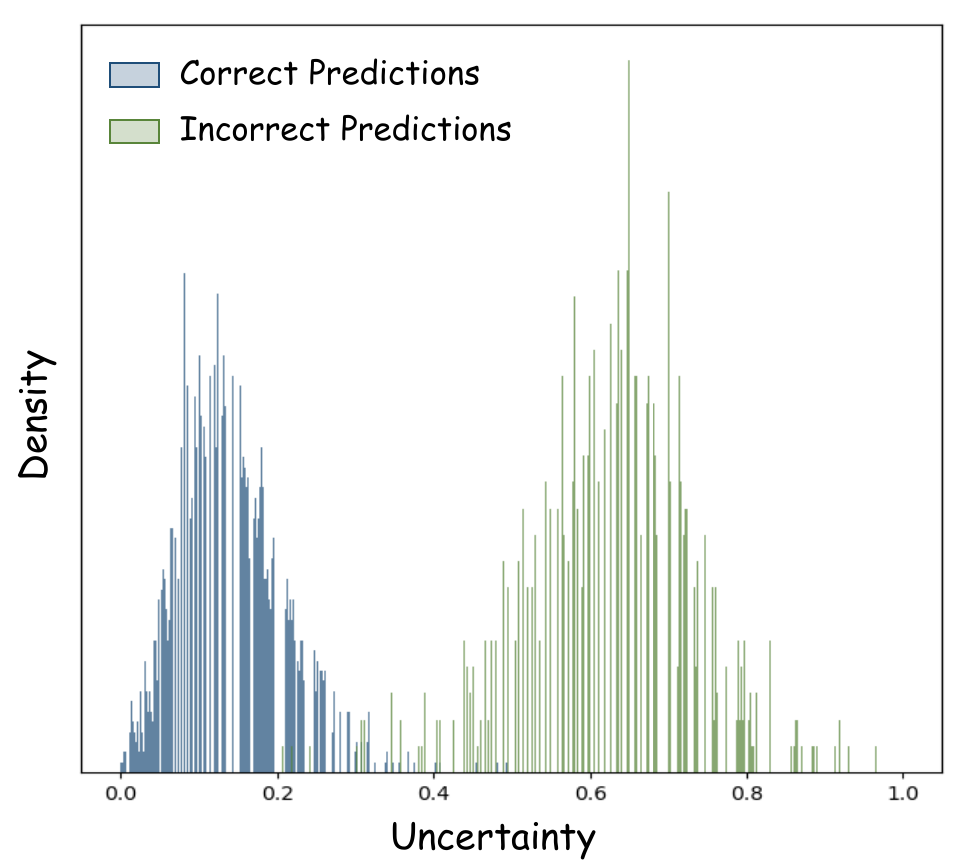}
\caption{Comparison of uncertainty distributions between correct and incorrect predictions.}
\label{dist}
\end{figure}

Ideally, the uncertainty estimated by the model should generally follow (though not strictly adhere to) the rule that \textbf{"the more accurate the prediction, the lower the uncertainty"}. In the experimental section of the main text, we have validated this rule by quantifying prediction accuracy using loss $\mathcal{L}_{\mathrm{v\overline{q}a}}$. Here, we revisit this point by comparing the uncertainty distributions of correct and incorrect predictions made by the model on the STAR validation set, as shown in Figure~\ref{dist}.

\subsection{Adversarial Experiments.}

In the methodology section of the main text, we assume that \textbf{”the model will regard low-quality video representations caused by insufficient modality alignment and low-quality questions as OOD context and will output higher uncertainty when answering"}. Here, we simulate low-quality video representations and low-quality questions by applying varying levels of Gaussian noise to the video features and by zeroing out different proportions of the question text respectively, on the STAR validation set, and examine the resulting uncertainty in the answers.

\begin{figure}[H]
\centering
\includegraphics[width=0.45\columnwidth]{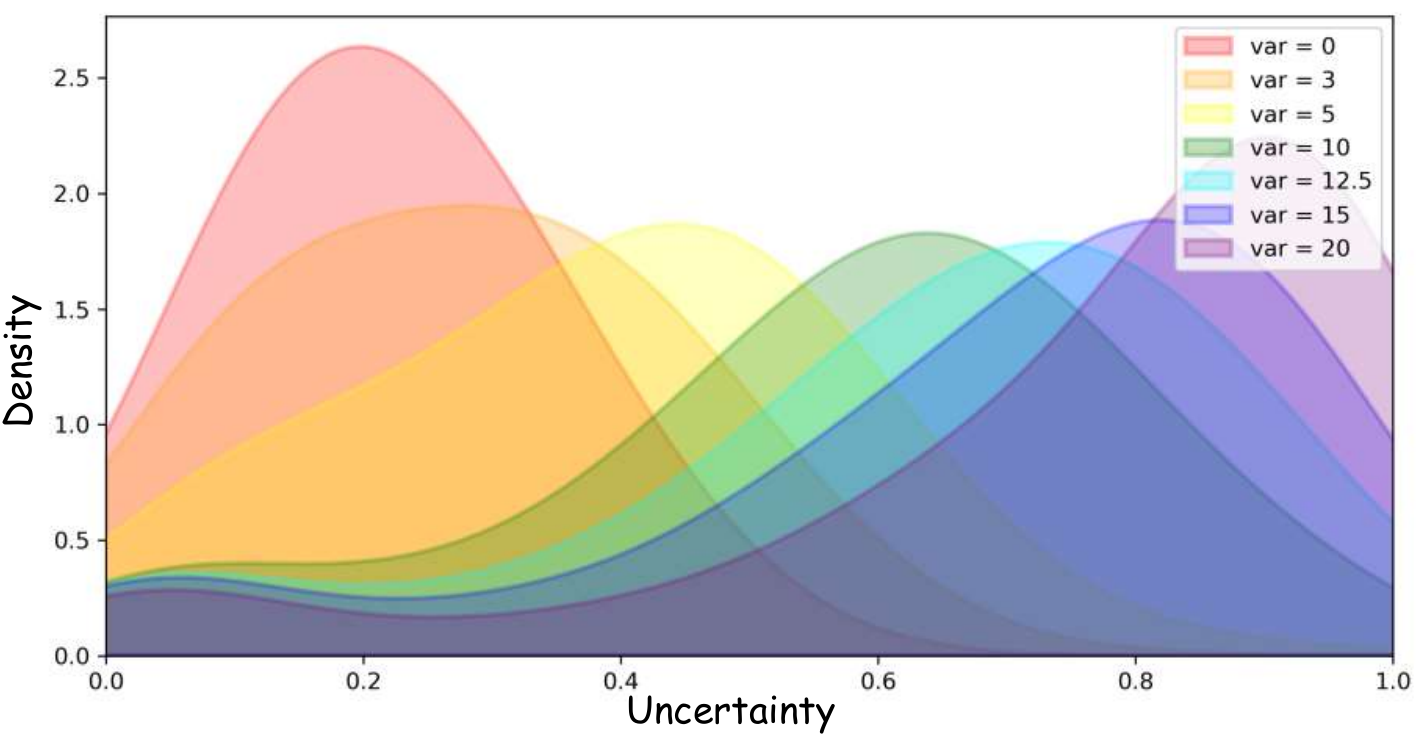}
\hspace{0.05\columnwidth}
\includegraphics[width=0.45\columnwidth]{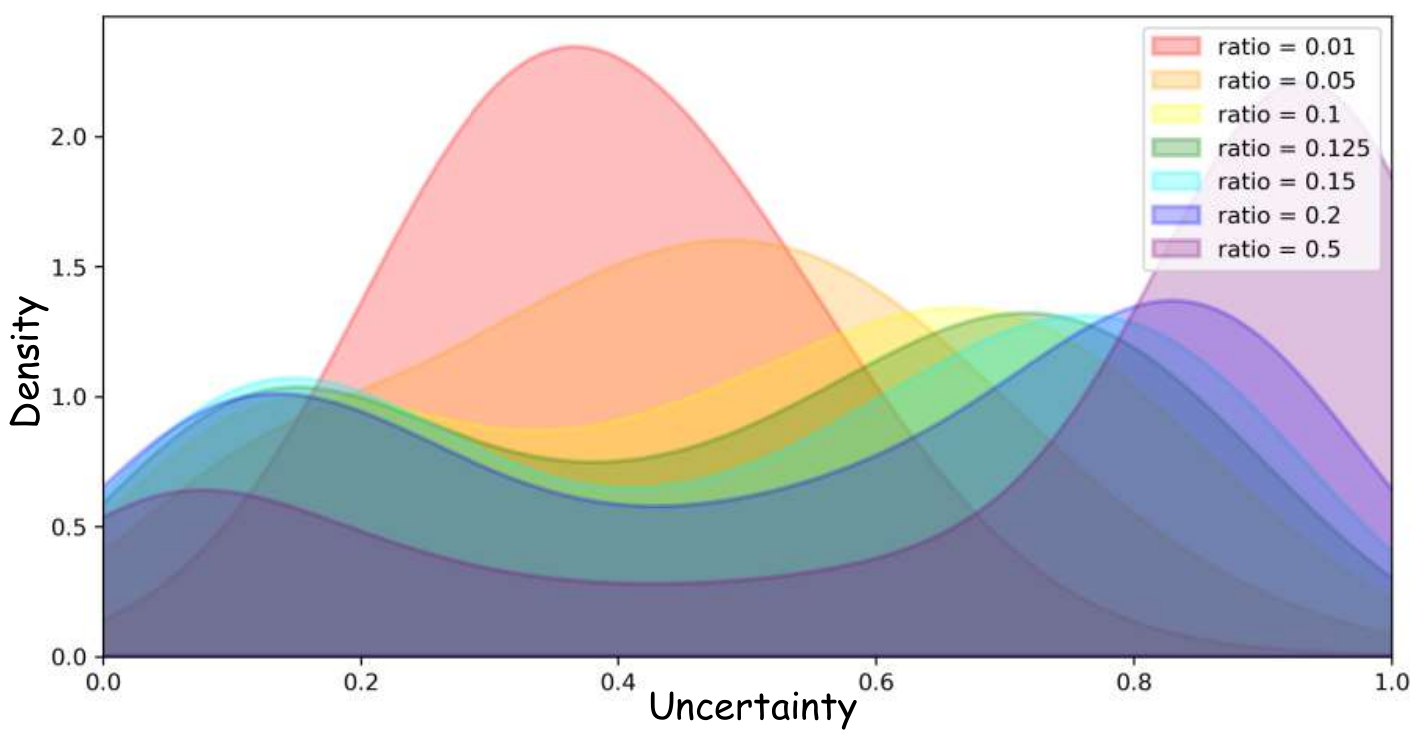}
\caption{\textbf{Left}: Comparison of the uncertainty distributions under different levels of video noise.
\textbf{Right}: Comparison of the uncertainty distributions under different levels of question destruction.}
\label{noise}
\end{figure}

We present a comparison of the uncertainty distributions when video features are destroyed in Figure~\ref{noise} left and textual features are destroyed in Figure~\ref{noise} right. It can be observed that as the degree of destruction increases, the uncertainty distribution generally tends to shift progressively to the right, which to some extent validates the hypothesis \textbf{”the model will regard low-quality video representations caused by insufficient modality alignment and low-quality questions as OOD context and will output higher uncertainty when answering"}.


\bibliographystyle{elsarticle-harv} 
\bibliography{ref}

\end{document}